\documentclass{article}
\usepackage{ijcai25}

\usepackage{times}
\usepackage{soul}
\usepackage{url}
\usepackage[hidelinks]{hyperref}
\usepackage[utf8]{inputenc}
\usepackage[small]{caption}
\usepackage{graphicx}
\usepackage{amsmath}
\usepackage{amsthm}
\usepackage{booktabs}
\usepackage{algorithm}
\usepackage{algorithmic}
\usepackage[switch]{lineno}

\title{Representation Learning with Mutual Influence of Modalities for Node Classification in Multi-Modal Heterogeneous Networks}

\author{
Jiafan Li$^{1,2}$
\and
Jiaqi Zhu$^{1,2,4}$\footnote{Corresponding author}
\and
Liang Chang$^3$\
\and
Yilin Li$^{1,2}$\
\and
Miaomiao Li$^{1,2}$\
\and
Yang Wang$^{1,2}$\
\and \\
Yi Yang$^1$\And
Hongan Wang$^{1,2}$\\
\affiliations
$^1$Institute of Software, Chinese Academy of Sciences, Beijing, China\\
$^2$University of Chinese Academy of Sciences, Beijing, China\\
$^3$School of Artificial Intelligence, Beijing Normal University, Beijing, China\\
$^4$Binzhou Institute of Technology, Weiqiao-UCAS Science and Technology Park, Shandong, China
\emails
\{lijiafan23, limiaomiao22, wangyang223\}@mails.ucas.ac.cn,
zhujq@ios.ac.cn, \\ \{yilin, yangyi2012, hongan\}@iscas.ac.cn,
changliang@bnu.edu.cn
}

\usepackage{multirow}
\usepackage{subfigure}
\usepackage{diagbox}

\newtheorem{definition}{Definition}

\usepackage{amssymb}

\begin{document}

\maketitle

\begin{abstract}
Nowadays, numerous online platforms can be described as multi-modal heterogeneous networks (MMHNs), such as Douban's movie networks and Amazon's product review networks. Accurately categorizing nodes within these networks is crucial for analyzing the corresponding entities, which requires effective representation learning on nodes. However, existing multi-modal fusion methods often adopt either early fusion strategies which may lose the unique characteristics of individual modalities, or late fusion approaches overlooking the cross-modal guidance in GNN-based information propagation. In this paper, we propose a novel model for node classification in MMHNs, named Heterogeneous Graph Neural Network with Inter-Modal Attention (HGNN-IMA). It learns node representations by capturing the mutual influence of multiple modalities during the information propagation process, within the framework of heterogeneous graph transformer. Specifically, a nested inter-modal attention mechanism is integrated into the inter-node attention to achieve adaptive multi-modal fusion, and modality alignment is also taken into account to encourage the propagation among nodes with consistent similarities across all modalities. Moreover, an attention loss is augmented to mitigate the impact of missing modalities. Extensive experiments validate the superiority of the model in the node classification task, providing an innovative view to handle multi-modal data, especially when accompanied with network structures.
\end{abstract}








\section{Introduction}

\begin{figure}
    \centering
    \includegraphics[width = 1\columnwidth]{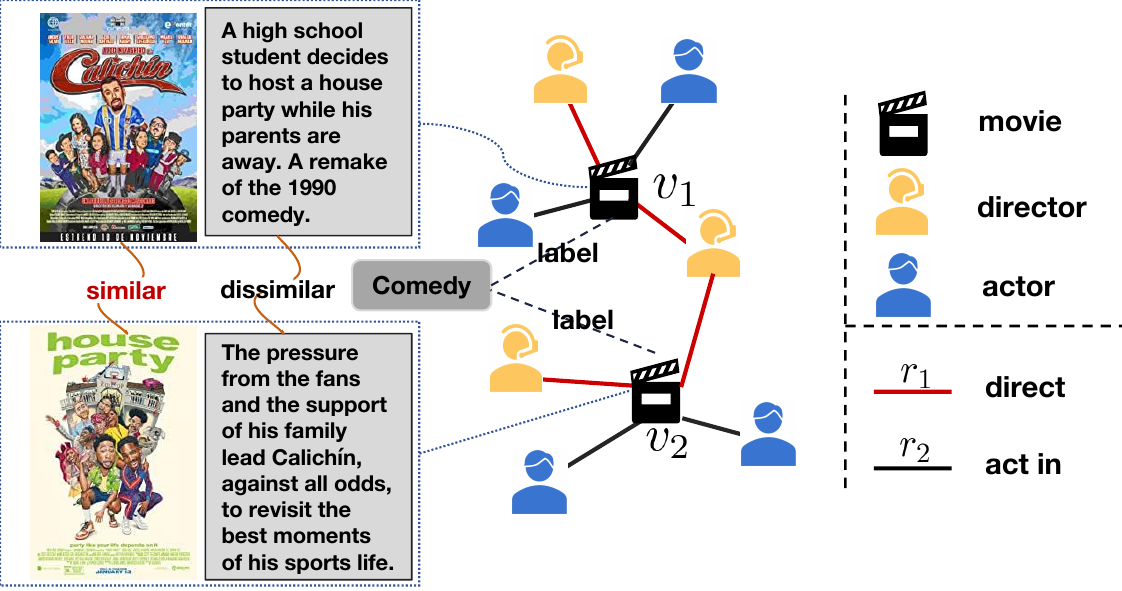}
    \caption{Motivating example of MMHNs and mutual influence of modalities in information propagation for node classification.}
    \label{fig:motivation}
\end{figure}

In the real world, numerous online platforms can be characterized as heterogeneous networks \cite{sun2012HIN}, which encompass multiple types of nodes connected with various relations, such as movie networks in Douban and product review networks in Amazon \cite{ni2019justifying}. With the rapid evolution of Internet, besides textual contents, nodes of certain types also incorporate attributes from other modalities (e.g., images). That enriches the node information and forms multi-modal heterogeneous networks (MMHNs) \cite{wei2023multimodal,kim2023heterogeneous,chen2020hgmf,mhgat}, including multi-modal knowledge graphs (MMKGs) \cite{kannan2020multimodal,liang2022KG,zhu2024MMKG,wang2023TIVAKG,mmkg,pezeshkpour2018embedding}. As shown in Figure~\ref{fig:motivation}, a movie network comprises movies, actors and directors, with the movie nodes possessing both textual and visual attributes.




It is significant to classify these nodes in the network into distinct categories to understand and analyze key entities, leveraging their multi-modal attributes as well as the network structure \cite{jangra2023summarizationsurvey}.
Recently, more and more attentions have been paid to integrate multi-modal features into GNN-based representation learning methods. Some existing studies adopt an early fusion strategy in the encoding process aiming at initial features
\cite{zhang2019HetGNN},
which may lose the characteristics of individual modalities, while late fusion models learn node embeddings separately for each modality and blend them only at the last layer 
\cite{mhgat}.

For example in Figure~\ref{fig:motivation}, the two film nodes $v_1$ and $v_2$ belong to the same category ``Comedy'', but their textual descriptions differ greatly. Hence, when learning textual embeddings aggregated from neighbors, the similarity-based attentive (2-hop) propagation between them would be limited, potentially resulting in the assignments of them to different categories. Nevertheless, noticing that the images of them are very similar, if that can be acknowledged in determining the propagation weights, the two nodes may eventually own similar embeddings on both modalities and obtain the same label. 
While, maybe in another scenario, textual similarity plays a pivotal role in realizing a category-oriented aggregation.



Consequently, in order to harness the multi-modal information more effectively in representation learning on MMHNs, it is essential to \textbf{consider the mutual influence of modalities during the information propagation process and learn it in an adaptive way}. This poses three main challenges: Firstly, as the influence involves at least two modalities and two nodes, 
we need to \textbf{choose the appropriate  granularity to define and distinguish the cross-modal influence}. Here, a tradeoff among expressiveness, intuitive interpretability and model complexity is supposed to be achieved. 
Secondly, in real-world scenarios, certain types of nodes may have \textbf{missing attributes for specific modalities} (e.g., actors and directors lack images), so it is required to handle these incomplete data to ensure a smooth feature propagation process for each modality. 
Thirdly, even for nodes with attributes across all modalities, some of these attributes may fail to accurately reflect node characteristics. This \textbf{misalignment among modalities} would import undesired noise during propagation, causing the learned representations to deviate from correct labels.



To address these issues, this paper proposes a novel model named Heterogeneous Graph Neural Network with Inter-Modal Attention (HGNN-IMA). It aims to learn node representations in MMHNs via capturing the mutual influence of multiple modalities during the information propagation process, thereby supporting the node classification task.
Specifically, after pre-processing to encode the attributes of each modality, we devise a heterogeneous network propagation module within the framework of heterogeneous graph transformer, to enrich the node features of each modality by aggregating neighbor information.
A key innovation is the nested inter-modal attention mechanism integrated into the classical inter-node attention. When propagating neighbor embeddings to a current node for a specific modality, attention scores are computed as the weighted sum of the similarity-based attention in terms of each modality. The weights are also determined in an attentive and thus adaptive manner, which are further constrained by an attention loss to mitigate the effect of missing modalities. Moreover, these attention scores are necessarily modulated according to the similarity consistency among modalities, amplifying the contribution of modality-aligned nodes. 
Then, through a feature fusion module, enriched features of all modalities are mixed to generate the final node embeddings.
Besides the cross-entropy loss on the fused features, uni-modal features are also incorporated to underline the respective effect of each modality.

In summary, this paper makes the following contributions:
\begin{itemize}
    \item 
    To the best of our knowledge, this is the first work to adaptively learn and leverage the mutual influence of multiple modalities for model fusion in GNN-based representation learning on MMHNs, tailored for the node classification task.
    \item 
    A novel model, framed within the heterogeneous graph transformer architecture, is proposed to fulfill the core idea. It features a nested inter-modal attention mechanism on the inter-node attention,
    plus a modulation term based on similarity consistency to encourage modality alignment, and an additional loss function tackling the modality missing issue. This comprehensive approach accommodates the intricate nature of cross-modal interactions in heterogeneous networks and enhances the category discriminability of node representations.
    \item 
    Extensive experiments on diverse real-world benchmark datasets demonstrate the effectiveness and stability of the model, achieving significant performance improvements over existing approaches for node classification.
\end{itemize}



\section{Related Work}



Heterogeneous networks widely exist in the real world.
In the past decade, significant efforts have been devoted to learn node representations with various attention mechanism \cite{zhuo2023propagation,wang2023CSA,he2023PSHGCN,li2023THGNN}.
For heterogeneous networks with multi-modal attributes, such as texts, images and audios, the fusion of them is necessary. Most of existing models fall into two types, early fusion and late fusion \cite{jangra2023summarizationsurvey,fusionsurvey,missingsurvey,tmlp}.

\textbf{Early fusion} means the fusion is conducted just after the feature extraction from multi-modal attributes \cite {wang2020multimodal,chen2020hgmf}. Besides extending the typical heterogeneous network embedding methods mentioned above by combining initial attributes,
HetGNN \cite{zhang2019HetGNN} employs a Bi-LSTM model to encode the feature of each modality and fuse them with mean pooling before intra-type and inter-type aggregations. 
As this kind of fusion appears prior to information propagation, all modalities of a specific node are merged into one feature, at the expense of losing the characteristics of individual modalities, especially neglecting their different roles in the aggregation process.

Conversely, \textbf{late fusion} refers to the fusion after the node representations in terms of each modality have been obtained, so the information propagation of each modality is separately executed \cite{wei2019mmgcn,tao2020mgat,cai2024multimodal,cao2022CKGC}. As representatives, MHGAT \cite{mhgat} performs dual-level aggregations within individual modalities, followed by feature fusion using the modality-level attention mechanism, and 
FHIANE \cite{FHIANE} adds an early fusion module, taking advantage of the consistency and complementarity of multi-modal information.
Although these models allow each modality to propagate independently, preserving its original properties, they fail to leverage the information from other modalities when computing the similarity-based attention for accurate aggregation weights. As a result, the learned embeddings may be inconsistent with category labels. 

In addition to the two manners of multi-modal fusion above, 
XGEA \cite{xu2023cross} considers the influence between two modalities during the propagation process, but the role of the current modality itself is neglected and the characteristics of each node in the propagation cannot be distinguished. In contrast, our model explores to adaptively learn and combine the mutual influence among modalities for each node, through a novel nested attention mechanism to enable flexible aggregations.

Beyond that, IDKG \cite{li2023incorporating} regards the knowledge graph as another modality. Its translation-based embeddings, along with visual and textual features, are fused into a unified representation, 
so is able to solve the node classification problem with a different usage of network structure.

\begin{figure*}
    \centering
    \includegraphics[width=1\textwidth]{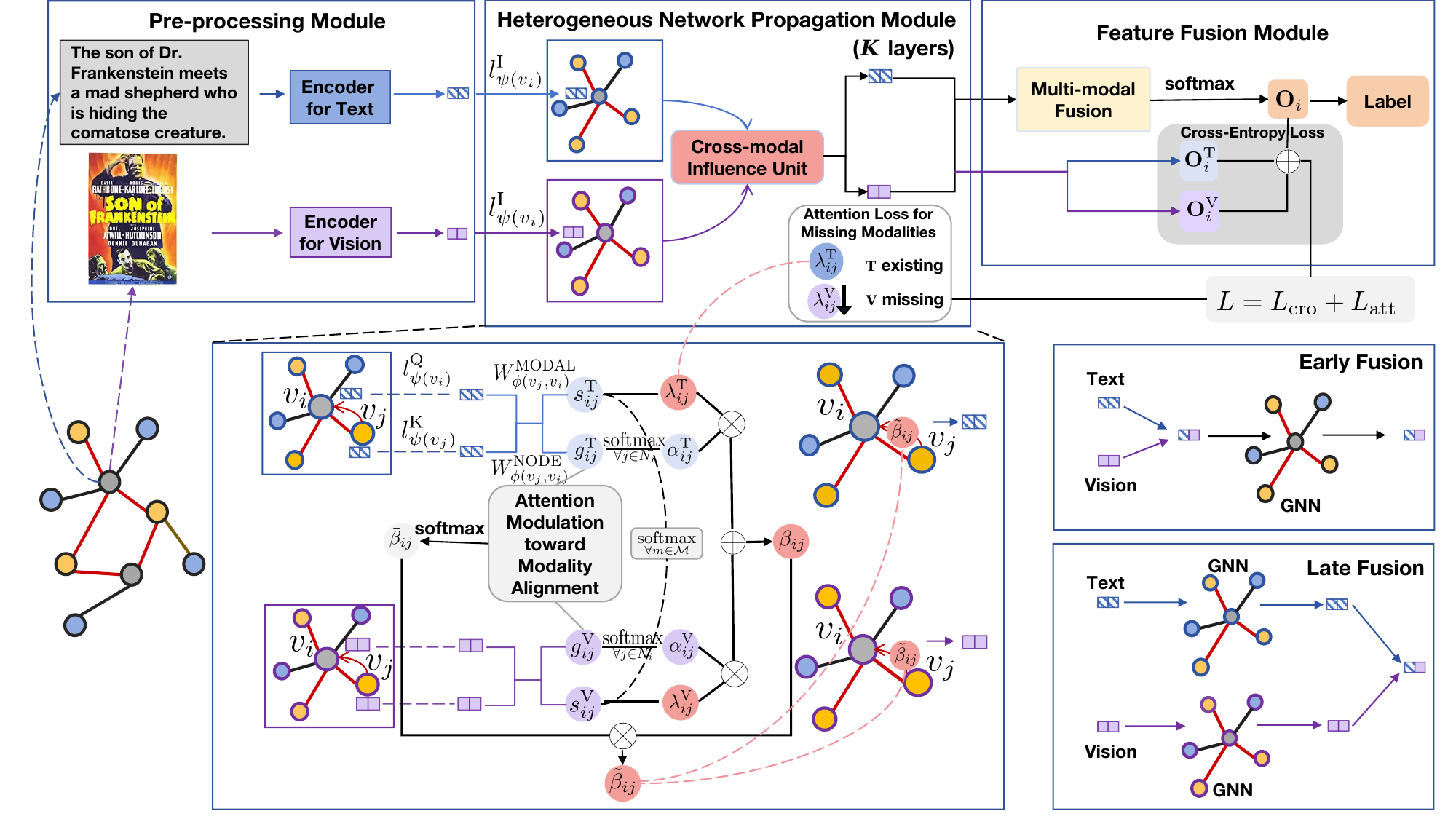}
    \caption{Framework of HGNN-IMA comprising of three modules, compared with existing modal fusion strategies on the bottom right.}
    \label{fig:frame}
\end{figure*}

\section{Problem Formulation}
\label{sec:formulation}

\begin{definition}[Multi-Modal Heterogeneous Networks]
\label{def:MMHN}
A Multi-Modal Heterogeneous Network (MMHN)  is defined as $ \mathcal G=( \mathcal V, \mathcal E, \mathcal M)$, where $ \mathcal V$, $ \mathcal E$ and $ \mathcal M$ respectively denote the set of nodes, edges and multi-modal attributes (e.g., texts, images and audios). 
There are also a node type mapping function $\psi(v): \mathcal V \to \mathcal O$ and an edge type mapping function $\phi(e): \mathcal E \to \mathcal R$, where 
$\mathcal O$ and $\mathcal R$ denote the set of pre-defined types of nodes and edges respectively, with
$\left|\mathcal O\right| + \left|\mathcal R\right| > 2$. 

Notice that although each node in $\mathcal V$ can contain multiple types of attribute information from different modalities in $\mathcal M$., some modalities are not available for certain node types. For example, reviews in Amazon are probably not associated with an image. Thus, we define
a mapping function $f(o):\mathcal O \to 2^{\mathcal M}$
to indicate the subset of modalities that each node type possesses.
In this context, for each node $v_i\in \mathcal{V}$, its attribute of each modality $m\in f(\psi(v_i))$ is denoted as $x_i^m$.

It is reasonable to assume there is only one target node type $o^* \in \mathcal{O}$ required to be categorized, such as films in movie networks and items in product review networks. Given a pre-defined set of categories $\mathcal{C}$, the node classification task on MMHNs aims to assign a category label $y_i \in \mathcal{C}$ to each target node $v_i$ in a semi-supervised setting. 

\end{definition}

\section{Proposed Model HGNN-IMA}

In this section, we at first present the framework of the proposed model, and then elaborate on the key modules.

\subsection{Framework}

HGNN-IMA is structured into three modules: pre-processing module, heterogeneous network propagation module, and feature fusion module.
Since most of real-world MMHNs have only text and vision attributes, we illustrate our model in Figure~\ref{fig:frame} with these two modalities, i.e., $\mathcal{M} = \{\textrm{T},\textrm{V}\}$.

As pre-processing, the feature for each modality of each node is extracted through a corresponding pre-trained encoder.
Then, the core propagation module is designed to learn enriched representations of nodes in the framework of heterogeneous graph transformer (HGT) \cite{hu2020HeterogeneousGraphTransformer}. To capture synergetic effect of multi-modal attributes, we primarily innovate with a \textbf{Cross-modal Influence Unit}, which nests the inter-modal attention into modal-specific inter-node attentions, allowing the aggregation weights to be determined by all modalities together. 
Besides, an additional loss for inter-modal attention and a modulation on inter-node attention are employed during propagation to mitigate the impact of nodes with missing and misaligned modalities respectively.

At last, the feature fusion module incorporates the
representations from different modalities also in an attentive manner.
For semi-supervised training, the cross-entropy loss on both multi-modal and uni-modal features, as well as the attention loss are combined with equal proportions.

\subsection{Heterogeneous Network Propagation Module}


In order to aggregate the embeddings of heterogeneous neighbors for the semantic enrichment of each node, 
we adopt HGT as the base architecture instead of dual-level ones \cite{hu2019HeterogeneousGraphAttention,wang2019heterogeneous} to uniformly handle various types of nodes and edges with the help of multiple type-dependent feature projection functions. 
That facilitates the characterization of complicated interactions among heterogeneous entities and their multi-modal attributes. 



Considering the discrepancy among node types, for each node $v_i$ and its feature on a modality $m$, we initially make a linear function dependent on node type to get the embedding $\mathbf{h}_{i}^{(0),m}$ at the 0-th layer, as the starting point of aggregation.
\begin{equation}
\mathbf{h}_{i}^{(0),m} =l^\textrm{I}_{\psi(v_i)} (\bar{\mathbf{h}}_{i}^{(0),m})
\end{equation}
where $\bar{\mathbf{h}}_{i}^{(0),m}$ is the output of the $m$-encoder in the pre-processing phase.
Then, each node $v_i$ expects to receive information from its neighbor set, denoted as $N_i\subseteq \mathcal{V}$.

\subsubsection{Cross-modal Influence Unit}

Like previous work of multi-modal fusion in GNNs, the propagation here is still executed for each modality $m$ (e.g., text or vision) separately, but as exemplified in Figure~\ref{fig:motivation}, other modalities should also take effect in deciding the importance (weights) of neighbors.
Thus, we need to compute these weights from the perspective of each influencing modality based on the similarity of the features on that modality, and unequally combine them in an adaptive way.


Specifically, for each influencing modality $m'\in \mathcal{M}$, given
a current node $v_i$ and one of its neighbors $v_j \in \mathcal N_{i}$,
the importance of $v_j$ for $v_i$ at the $k$-th layer can be calculated as the similarity of the two nodes on this modality, and realized by a bilinear transformation following HGT \cite{hu2020HeterogeneousGraphTransformer}:
\begin{equation}
\label{g}
g_{ij}^{(k),m'} = l^\textrm{K}_{\psi(v_j)}(\mathbf{h}_j^{(k-1),m'}) \cdot W^\textrm{NODE}_{\phi(v_j,v_i)} \cdot l^\textrm{Q}_{\psi(v_i)}(\mathbf{h}_i^{(k-1),m'})
\end{equation}
where $l^\textrm{K}$ and $l^\textrm{Q}$ are two linear functions dependent on the node type similar to $l^\textrm{I}$, and $W^\textrm{NODE}$ is a learnable matrix dependent on the edge type. 
Then, the inter-node attention score between $v_i$ and its neighbor $v_j$
at the $k$-th layer on the modality $m'$ is denoted as $\alpha_{ij}^{(k),m'}$ and computed as follows:
\begin{equation}
\label{alpha}
\alpha_{ij}^{(k),m'} = \mathop{\textrm{softmax}}\limits_{\forall j\in N_i}\left(g_{ij}^{(k),m'} \right)=\frac{\exp\left(g_{ij}^{(k),m'}\right)}{\sum_{j' \in N_{i}} \exp\left(g_{ij'}^{(k),m'}\right)}
\end{equation}
Here, a single attention head is used for simplicity, and is easy to be extended to multiple heads similar to HGT.

With these attention scores on each modality, it is crucial to determine which modalities are suited to play more important roles. That naturally depends on the features of the two nodes $v_i$ and $v_j$, so should be learned adaptively. To this end, we design a nested attention mechanism. At first, the inter-modal attention score for the modality $m'$ is calculated in two steps:
\begin{equation}
\label{s}
s_{ij}^{(k),m'} = l^\textrm{K}_{\psi(v_j)}(\mathbf{h}_j^{(k-1),m'}) \cdot W^\textrm{MODAL}_{\phi(v_j,v_i)} \cdot l^\textrm{Q}_{\psi(v_i)}(\mathbf{h}_i^{(k-1),m'})
\end{equation}
\begin{equation}
\label{lambda}
\lambda_{ij}^{(k),m'} = \mathop{\textrm{softmax}}\limits_{\forall m'\in \mathcal{M}}\left(s_{ij}^{(k),m'} \right)=\frac{\exp\left(s_{ij}^{(k),m'}\right)}{\sum_{m''\in \mathcal{M}} \exp\left(s_{ij}^{(k),m''}\right)}
\end{equation}

Notice that we use the same features of the two nodes as the inter-node attention $\alpha$, but different parameters $W^\textrm{MODAL}$ to characterize their correlations from the perspective of cross-modal influence, instead of similarity. 


Next, we apply the inter-modal attention $\lambda_{ij}^{(k),m'}$ in Equation~\ref{lambda} on the modal-specific inter-node attention $\alpha_{ij}^{(k),m'}$ in Equation~\ref{alpha}, and obtain the combined inter-node attention $\beta_{ij}^{(k)}$ reconciling the features and effects of all modalities:
\begin{equation}
\label{beta}
    \beta_{ij}^{(k)} =  \mathop{\textrm{softmax}}\limits_{\forall j\in N_i} \left(\sum_{m'=1}^{\mathcal M} \left( \lambda_{ij}^{(k),m'} \alpha_{ij}^{(k),m'} \right) \right)
\end{equation}

It is worth noting that this  attention is independent of the current (influenced) modality $m$. In other words, for a specific node, the features of all modalities are propagated according to unique weights from its neighbors. Although compromising the expressiveness to some extent, this simplification is reasonable to maintain a moderate number of parameters and highlight the essential idea of the cross-modal influence, which is certified via ablation study in Section~\ref{sec:ablation}.

In view of the complexity of multi-modal data, there exist the phenomenons of modality misalignment and even missing. Therefore, the inter-modal attention and the inter-node attention need to be adjusted accordingly to fulfill a smooth and category-oriented propagation.

\subsubsection{Attention Loss for Missing Modalities}

As formulated in Section~\ref{sec:formulation}, there exist some types of nodes not possessing all modalities originally, and their features on the missing modalities have to be completed in some way, so they should not take much effect compared to modalities with real attributes.
To address this issue, we specially design a loss function to constrain the inter-modal attention scores computed by Equation~\ref{lambda} in such cases are not too large:
\begin{equation}
\label{Latt}
L_\textrm{att} =  \frac{1}{K \cdot |\mathcal{M}|} 
\sum_{v_i\in \mathcal{V}}~
\sum_{v_j\in N_i}~
\sum_{1\le k\le K}~
\sum_{m'\not\in f(\psi(v_j))}~\lambda_{ij}^{(k),m'}
\end{equation}
where $K$ is the number of propagation layers. 






\subsubsection{Attention Modulation toward Modality Alignment}

Even for those nodes possessing all modalities, the attributes of some modalities may not represent the correct meaning of nodes due to the irregularity of Internet. If a node suffering such misalignment propagates too much information to the current node, it is inevitable to import noises in the aggregation.
Since the similarity of two nodes on each modality has been computed in Equation~\ref{g}, we can utilize the consistency of these scores to imply the modality alignment degree of each neighbor, and thus get a new inter-node attention $\bar{\beta}$ for the aggregation weights as follows. 
\begin{equation}
    \bar{\beta}_{ij}^{(k)} = 
    \mathop{\textrm{softmax}}\limits_{\forall j\in N_i}\left(\sum_{m_1,m_2\in \mathcal{M}}|g_{ij}^{(k),m_1}-g_{ij}^{(k),m_2}|\right) 
\end{equation}

Then, through mixing the two inter-node attentions from different views into final aggregation weights $\tilde{\beta}_{ij}^{(k)}$, the embedding of each modality $m$ 
at the $k$-th layer is computed as:
\begin{equation}
\label{tildebeta}
\tilde{\beta}_{ij}^{(k)} =  \mathop{\textrm{softmax}}\limits_{\forall j\in N_i}(\beta_{ij}^{(k)} \cdot \bar{\beta}_{ij}^{(k)})
\end{equation}
\begin{equation}
\label{eq:aggregation}
\tilde{\mathbf{h}}_{i}^{(k),m} = \sum_{j \in N_{i}} \tilde{\beta}_{ij}^{(k)} \cdot l^\textrm{M}_{\psi(v_j)}(\mathbf{h}_{j}^{(k-1),m}) \cdot W^\textrm{MSG}_{\phi(v_j,v_i)} 
\end{equation}
where $l^\textrm{M}$ is the fourth function dependent on node type and $W^\textrm{MSG}$ is the third learnable matrix dependent on edge type.

At last, to avoid over-smoothing, we introduce a residual connection to get
the final output at the $k$-th layer, with the sigmoid function and another type-dependent function $l^\textrm{A}$:
\begin{equation}
\mathbf{h}_{i}^{(k),m} = l^\textrm{A}_{\psi(v_i)}\left(\sigma(\tilde{\mathbf{h}}_{i}^{(k),m})\right) + \mathbf h_{i}^{(k-1),m} 
\end{equation}

\subsection{Feature Fusion Module}

Although the computation of embeddings for each modality has already taken cross-modal influence into accounts, they are still required to be fused to get final representations. 
We use standard modality-level attention to learn adaptive importance of each modality for classification at the last layer:
\begin{equation}
    \omega_i^{m} = \mathbf{w_2} \cdot \tanh(\mathbf{w_1} \cdot \mathbf{h}_i^{(K),m}) + b_2
\end{equation}
\begin{equation}
  \delta_{i}^{m} = \text{softmax}(\omega_{i}^{m}) = \frac{\exp(\omega_{i}^{m})}{\sum_{m'' \in \mathcal M} \exp(\omega_{i}^{m''})}  
\end{equation}
The final embedding $\mathbf{Z}_{i}$ of node $v_i$
is then calculated, and the probability distribution $\mathbf{O}_i$ of $v_i$ for each category is obtained:
\begin{equation}
\mathbf{Z}_{i} = \sum_{m \in \mathcal M} \delta_{i}^{m} \cdot \mathbf{h}_{i}^{(K),m} 
\end{equation}
\begin{equation}
\label{eq:output}
    \mathbf{O}_i = \textrm{softmax}(\mathbf{W}_1 \cdot \mathbf{Z}_i)
\end{equation}

\subsection{Training Objective}

For semi-supervised classification,
the cross-entropy loss is used. 
To embody the effect of each modality, we incorporate the losses of individual modalities into the fused one: 
\begin{equation}
L_{\text{cro}} =  \frac {1}{1+|\mathcal M|}( \sum_{v_i \in \mathcal{V}^*_\mathrm{L}} \mathbf{y}_i^\top \cdot \log(\mathbf{O}_{i}) + \sum_{m \in \mathcal M}\sum_{v_i \in \mathcal{V}^*_\mathrm{L}} \mathbf{y}_i^\top \cdot \log(\mathbf{O}_{i}^m))
\end{equation}
where $\mathcal{V}^*_\mathrm{L}$ is the labeled target node set, and $\mathbf{y}_i$ is the one-hot label vector for $v_i$.
$\mathbf{O}_{i}^m$ is the embedding for each modality $m$, computed similar to  $\mathbf{O}_i$ in Equation~\ref{eq:output}, but replacing $\mathbf{Z}_i$ by $\mathbf{h}_{i}^{(K),m}$. 
Then, the whole loss function is expressed as the equal-weight sum of the two losses:
\begin{equation}
    L=L_\textrm{cro}+L_\textrm{att}
\end{equation}
It can be inferred that the total computational complexity is $O(|\mathcal{V}|^2 \cdot |\mathcal{M}|^2)$,
which demonstrates the scalability of the model when handling large graphs and multiple modalities. To save space, the detailed analysis is put in Appendix \ref{sec:complexity}.



\section{Experiments}

In this section, we first introduce the datasets, baselines, and experimental settings. Subsequently, we demonstrate the superiority of our model over baselines on node classification, followed by ablation study and hyper-parameter analysis.

The experiments were performed on NVIDIA Tesla V100 32 GPUs, and implemented in Python 3.9 with PyTorch.\footnote{The code is available at \url{https://github.com/Jiafan-ucas/HGNN-IMA}}

\subsection{Datasets}
We evaluate our model on five diversified real-world benchmark datasets. The statistics of them are shown in Table~\ref{tab:statistic}.
\begin{itemize}
    \item \texttt{DOUBAN}\footnote{https://github.com/jiaxiangen/MHGAT/tree/main/douban} and \texttt{IMDB}\footnote{https://github.com/Jhy1993/HAN/tree/master/data/imdb} collect data from two online movie websites respectively. Movies (M), actors (A) and directors (D) compose the heterogeneous network with edge types AM, MA, MD and DM.
    Besides textual descriptions, each movie possesses a poster image, and can be categorized into ``Action'', ``Comedy'' and ``Drama'' (``Drama'' only exists in \texttt{IMDB}). 
    \item \texttt{AMAZON}\footnote{https://github.com/jiaxiangen/MHGAT/tree/main/amazon} contains items (I) and user reviews (U) in Amazon,
    with three types of relations: UI, IU and II. 
    Each item (product) has a text and an image. 
    Three sub-categories of appliances are used for classification.
\item 
\texttt{AMAZON-1} and \texttt{AMAZON-2} are self-constructed larger datasets under the Electronics category in the AMAZON dataset\footnote{https://nijianmo.github.io/amazon/index.html}.
Four types of relation between items are selected: also\_buy, also\_viewed, buy\_after\_viewing, and bought\_together.
In \texttt{AMAZON-2}, the price of items is added as the third modality.
   
\end{itemize}  
\begin{table}[]
\small
\centering
\begin{tabular}{ccccc}
\toprule
Dataset & Nodes & Edges  & Edge types & Categories
\\ \midrule
\texttt{DOUBAN}  & 6627  & 15032  & 4          & 2       \\ 
\texttt{IMDB}   & 11616 & 34212  & 4          & 3        \\  
\texttt{AMAZON}  & 13189 & 174154 & 3          & 3     \\ 
\texttt{AMAZON-1}  & 58088  & 632238  & 4          & 12        \\ 
\texttt{AMAZON-2}  & 58088  & 632238  & 4          & 12      \\ 
\bottomrule
\end{tabular}
  \caption{Dataset statistics.}
  \label{tab:statistic}
\end{table}



\begin{table*}
  \resizebox{\textwidth}{!}{
   \small

\small
\centering
\setlength{\tabcolsep}{5pt} 
\begin{tabular}{cl|cc|cc|cc|cc|cc}
\toprule
\multicolumn{2}{c|}{\multirow{2}{*}{\textbf{Datasets}}} & \multicolumn{2}{c|}{\multirow{2}{*}{\texttt{DOUBAN}}} & \multicolumn{2}{c|}{\multirow{2}{*}{\texttt{IMDB}}} & \multicolumn{2}{c|}{\multirow{2}{*}{\texttt{AMAZON}}} & \multicolumn{2}{c|}{\multirow{2}{*}{\texttt{AMAZON-1}}} & \multicolumn{2}{c}{\multirow{2}{*}{\texttt{AMAZON-2}}} \\
\multicolumn{2}{c|}{} & \multicolumn{2}{c|}{} & \multicolumn{2}{c|}{} & \multicolumn{2}{c|}{} & \multicolumn{2}{c|}{} & \multicolumn{2}{c}{} \\
\midrule
\multicolumn{2}{c|}{\textbf{Metrics}} & Micro-F1 & Macro-F1 & Micro-F1 & Macro-F1 & Micro-F1 & Macro-F1 & Micro-F1 & Macro-F1 & Micro-F1 & Macro-F1 \\
\midrule
\multicolumn{1}{c|}{\multirow{2}{*}{\textbf{HAN}}} & early & 0.8707 & 0.8666 & 0.7267 & 0.7262 & 0.8594 & 0.8015 & 0.8532 & 0.6866 & 0.8542 & 0.6927 \\
\multicolumn{1}{c|}{} & late & 0.8737 & 0.8699 & 0.7300 & 0.7286 & 0.8337 & 0.7737 & 0.8250 & 0.6157 & 0.8208 & 0.5936 \\
\midrule
\multicolumn{1}{c|}{\multirow{2}{*}{\textbf{SHGP}}} & early &0.8319 & 0.8288 & 0.5488 & 0.5447 & 0.7483& 0.6344 & 0.5989 &0.3311  &0.5678  &0.3038  \\
\multicolumn{1}{c|}{} & late & 0.8224 & 0.8256 & 0.5320 & 0.5180 & 0.7748& 0.6920 & 0.5911 & 0.3205 &0.5844  &0.3255  \\
\midrule
\multicolumn{1}{c|}{\multirow{2}{*}{\textbf{SeHGNN}}} & early & 0.8667 & 0.8652 & \textit{0.7496} & \textit{0.7478} & 0.8726 & 0.8289 & 0.8554 & 0.7323 & 0.8522 & 0.7561 \\
\multicolumn{1}{c|}{} & late & 0.8677 & 0.8624 & 0.7453 & 0.7438 & 0.8550 & 0.8122 & 0.8571 & 0.7638 & 0.8554 & 0.7660 \\
\midrule
\multicolumn{1}{c|}{\multirow{2}{*}{\textbf{HERO}}} & early & 0.8533 & 0.8493 & 0.6517 & 0.6102 & 0.8295 & 0.7699 & 0.8023 & 0.6755 & 0.8058 & 0.6546 \\
\multicolumn{1}{c|}{} & late & 0.8283 & 0.8252 & 0.6936 & 0.6848 & 0.8207 & 0.7547 & 0.8136 & 0.6862 & 0.8012 & 0.6723 \\
\midrule
\multicolumn{1}{c|}{\multirow{2}{*}{\textbf{HGT}}} & early & 0.8508 & 0.8483 & 0.7407 & 0.7381 & \textit{0.8773} & \textit{0.8302} & 0.8883 & 0.7682 & \textit{0.8882} & \textit{0.7807} \\
\multicolumn{1}{c|}{} & late & 0.8654 & 0.8629 & 0.7419 & 0.7407 & 0.8703 & 0.8212 & \textit{0.8931} & \textit{0.7990} & 0.8871 & 0.7799 \\
\midrule
\multicolumn{2}{c|}{\textbf{HetGNN} (early)} & 0.8366 & 0.8332 & 0.5068 & 0.4906 & 0.8328 & 0.7636 & 0.7012 & 0.5187 & 0.7129 & 0.4977 \\
\midrule
\multicolumn{1}{c|}{\multirow{2}{*}{\textbf{MHGAT} (late)}} & max & 0.8629 & 0.8574 & 0.7364 & 0.7249 & 0.8638 & 0.8084 & 0.8011 & 0.6729 & 0.8003 & 0.6486 \\
\multicolumn{1}{c|}{} & sum & {0.8545} & {0.8468} & {0.7220} & {0.7127} & {0.7963} & {0.6734} & {0.7975} & {0.5929} & {0.7873} & {0.5433} \\
\midrule
\multicolumn{2}{c|}{\textbf{IDKG}} & 0.8462 & 0.8451 & 0.7410 & 0.7387 & 0.8752 & 0.8276 & 0.8504 & 0.5164 & 0.8604 & 0.5268 \\
\midrule
\multicolumn{2}{c|}{\textbf{XGEA}} & \textit{0.8765} & \textit{0.8728} & 0.7126 & 0.7047 & 0.8596 & 0.8001 & 0.8847 & 0.7226 & 0.8872 & 0.7301 \\
\midrule
\multicolumn{2}{c|}{\textbf{HGNN-IMA}} & \textbf{0.8778} & \textbf{0.8758} & \textbf{0.7578} & \textbf{0.7560} & \textbf{0.8870} & \textbf{0.8427} & \textbf{0.8946} & \textbf{0.8233} & \textbf{0.8905} & \textbf{0.8182} \\
\bottomrule
\end{tabular}

}
  \caption{Overall results of HGNN-IMA and baselines on five datasets by two metrics. The best scores are in bold and the second in italic.}
  \label{tab:overall}
\end{table*}

\subsection{Baselines}
\label{sec:baselines}
For baselines, we choose representative models of the three manners regarding modalities: (1) typical heterogeneous graph neural networks without handling multi-modal attributes (\textbf{HAN} \cite{wang2019heterogeneous},  \textbf{SHGP}  \cite{yang2022selfsupervised}, \textbf{SeHGNN} \cite{sehgnn}, \textbf{HERO} \cite{mo2024HERO}) and \textbf{HGT} \cite{hu2020HeterogeneousGraphTransformer}, for which we employ \textit{both early fusion and late fusion} strategies;
(2) special heterogeneous graph neural networks tackling multi-modal attributes 
(\textbf{HetGNN} \cite{zhang2019HetGNN} by \textit{early fusion} , \textbf{MHGAT} \cite{mhgat} by \textit{late fusion} with two versions of inter-node aggregation, and \textbf{XGEA} \cite{xu2023cross} considering \textit{fixed influence between modalities}); (3) translation-based methods treating the network structure a new modality (\textbf{IDKG} \cite{li2023incorporating}).
The details are provided in Appendix \ref{sec:baseline-detail}.

\subsection{Experimental Settings}

As to \texttt{AMAZON}, \texttt{IMDB} and \texttt{DOUBAN} datasets, we directly use the encoded features  provided by MHGAT \cite{mhgat}. For self-constructed datasets \texttt{AMAZON-1} and \texttt{AMAZON-2}, texts and images are encoded using CLIP \cite{clip}, while price is embedded through an FFNN.
For missing visual attributes, we complete them with the text features after encoding, as the only available information.

We divide each dataset into the training set (20\%), the validation set (10\%), and the test set (70\%). The model is trained
using the Adam optimizer with a learning rate of 0.001, incorporating 3 layers of propagation ($K=3$) and setting the maximum number of iterations as 300. Each layer's output is subjected to a dropout with a rate of 0.6, and the dimension $d$ of all embeddings is standardized to 64. 
During propagation, 
we employ multi-head attention with the number of heads set to 8. To further prevent over-fitting, we employ an early stopping mechanism with a patience of 50, which activates when the validation loss exceeds any previously recorded values and its accuracy dips below the highest recorded one.
Both Micro-F1 and Macro-F1 are used for evaluation. We report the average results of five executions with different seeds.

\subsection{Overall Results}


Table \ref{tab:overall} shows the overall results on five datasets. 
It can be seen that our model consistently outperforms all the baselines across all datasets.
For the first three datasets, there are 1.2\%, 1.6\%, and 0.3\% gains in Macro-F1 compared to the strongest baseline. While, for the two larger datasets with more categories, the Macro-F1 value is promoted by around 2.5\%. 
That indicates the model is able to achieve a more balanced performance improvement across all categories on larger datasets.

Specifically, HGNN-IMA is superior to typical HGNNs such as HAN, SHGP, HERO and SeHGNN to a great extent, no matter early pre-processing or late post-processing for multiple modalities, so the special treatment on multi-modal attributes in MMHNs is necessary to understand the categories of entities and thereby worth studying. The second-best performances of HGT on most datasets are attributed to its use of type-dependent parameters to capture heterogeneous attentions over each edge from a global view. Our model further enhances HGT via effectively handling modalities.

Compared to HetGNN with explicit early concatenation of multi-modal attributes, as well as MHGAT employing late modality-level attention mechanism, the notable promotion of our model can naturally give credit to the innovative nested attention mechanism. It adaptively learns the cross-modal influence and determines the aggregation weights of each node to attain expected information propagation, rather than just blending multi-modal features before or after the propagation. 
Although XGEA considers the influence of another modality in propagation, it assumes a fixed influence relation and neglects the role of the current modality itself, so fails to identify the complicated synergy of modalities on category-oriented propagation for each node. 
Also, the advantage over IDKG
implies GNN-based propagation is essential to learn discriminative node embeddings through incorporating high-order correlations, 
especially facing multi-modal attributes.



Besides, we calculate the standard deviation (STD) of our model and sub-optimal HGT, and conduct a two-sample t-test between them, 
presented in Table \ref{tab:std}. The results confirm the stability of our model. Additionally, the p-value is less than 0.05 in all datasets, indicating a significant difference between HGNN-IMA and baseline models.
The intuitive comparison on learned embeddings is displayed in Appendix \ref{sec:visual-embedding}.
 
\begin{table}[]
 \small
 \centering
\setlength{\tabcolsep}{2pt} 
\begin{tabular}{cccccc}
\toprule
\textbf{Metrics}       & \texttt{DOUBAN} & \texttt{IMDB} & \texttt{AMAZON} & \texttt{AMAZON-1} & \texttt{AMAZON-2} \\ 
\midrule
\textbf{Ours (STD)} & 0.0015          & 0.0031        & 0.0019          & 0.0021            & 0.0033            \\ 
\textbf{HGT (STD)}      & 0.1261          & 0.0035        & 0.0072          & 0.0106            & 0.0076            \\ 
\textbf{t-value}        & 3.01            & 4.18          & 4.09            & 4.36              & 6.14              \\ 
\textbf{p-value}        & 0.039           & 0.014         & 0.015           & 0.012             & 0.004             \\ 
\bottomrule
\end{tabular}
 \caption{Standard deviation and t-test for HGNN-IMA and HGT.}
 \label{tab:std}
\end{table}

\subsection{Ablation Study}
\label{sec:ablation}
Here, we design hierarchical variants of HGNN-IMA to inspect the importance of the key components in the model. Other ablative models are analyzed in Appendix \ref{sec:ablation2}.
\begin{itemize}
    \item 
    Removing or changing the Cross-modal Influence Unit
    \begin{itemize}
        \item HGNN-IMA$-$cross: It eliminates this unit and directly uses inter-node attention $\alpha_{ij}^{(k),m'}$ to replace $\beta_{ij}^{(k)}$in Equation \ref{beta} for the aggregation.

                \item HGNN-IMA$-$adapt: It utilizes the mean of $\alpha_{ij}^{(k),m'}$ on each modality to serve as $\beta_{ij}^{(k)}$ in Equation \ref{beta}, losing weight adaptability.
                
        \item  HGNN-IMA$+$inf: It distinguishes the influenced modality $m$ when computing the inter-modal attention to form $\lambda_{ij}^{(k),m',m}$ in Equation \ref{lambda}.
        \item HGNN-IMA$-$nei: It neglects the distinction among the neighbors $v_j$ when computing the inter-modal attention to form $\lambda_{i}^{(k),m'}$  in Equation \ref{lambda}.
    
          \end{itemize}
    \item 
    Removing attention modulation for modality alignment
    \begin{itemize}

        \item HGNN-IMA$-$align: It directly uses inter-node attention $\beta_{ij}^{(k)}$ to replace $\tilde{\beta}_{ij}^{(k)}$ in Equation \ref{eq:aggregation}. 
    
    \end{itemize}
    \item Removing some part of loss functions
    \begin{itemize}
        \item   HGNN-IMA$-L_{\rm{att}}$: It removes the attention loss to disregard the modality missing issue.
      \item HGNN-IMA$-L_{\rm{ind}}$: It removes the individual modality part from the cross-entropy loss.
    \end{itemize}
        
\end{itemize}

Table \ref{tab:ablation} proves that the Cross-modal Influence Unit is beneficial,
and the weights of influence should be adaptively learned rather than pre-defined.
We also find compared to the traditional attention score $\alpha_{ij}^{(k),m}$,
our nested one $\beta_{ij}^{(k)}$ aligns more closely with node categories, which is visualized in Appendix \ref{sec:visual-attention}. Regarding the specific design of the nested inter-modal attention, when altering the granularity such as adding the influenced modality or removing the influencing node, the performance declines in varying degrees. Hence, while the core idea is simple, the tradeoff between expressiveness and complexity in defining the cross-modal influence is nuanced. 

Then, when removing the modulation term, the model exhibits decreased performance for all datasets. That certifies the unconstrained propagation according to the cross-modal influence is susceptible to noises imported by inaccurate modality attributes.
The modulation based on similarity consistency just weakens this awkward impact through focusing more on those nodes which realize modality alignment.

At last, the ablation for loss functions highlights their respective roles within the model.
An exception appears when the single-modality loss is removed in the \texttt{AMAZON-2} dataset. That can be explained as the introduction of the new modality, so it should be cautious to consider the contribution of individual modalities, which may bring a negative impact.

\begin{table}
   \small
\setlength{\tabcolsep}{2pt} 
\centering
\begin{tabular}{cccccc}
\toprule
\textbf{Variants} & \texttt{DOUBAN} & \texttt{IMDB} & \texttt{AMAZON} & \texttt{AMAZON-1} & \texttt{AMAZON-2} \\ 
\midrule
-cross          & \textit{0.8661}          & 0.7344        & 0.8173          & 0.8125          & 0.8067          \\
-adapt  & 0.8594          & 0.7360        & \textit{0.8397}          & 0.8044          & 0.7696          \\
+inf            & 0.8614          & 0.7319        & 0.8332          & 0.8101          & 0.7798          \\
-nei            & 0.8599          & 0.7256        & 0.8241          & 0.8082          & 0.7921          \\ 
\midrule
-align          & 0.8562          & \textit{0.7487}        & 0.8389          & 0.8056          & 0.8095          \\ 
\midrule
-$L_{\rm{att}}$ & 0.8467          & 0.7436        & 0.8334          & \textbackslash  & \textbackslash  \\ 
-$L_{\rm{ind}}$ & 0.8602          & 0.7469        & {0.8392}          & \textit{0.8218}          & \textbf{0.8270} \\ 
\midrule
               \textbf{Ours} & \textbf{0.8758} & \textbf{0.7560} & \textbf{0.8427} & \textbf{0.8233} & \textit{0.8182}          \\ 
\bottomrule
\end{tabular}
  \caption{Macro-F1 of ablative models. The best scores are marked in bold and second in italic. ``$\backslash$'' means inapplicable due to full data.}
  \label{tab:ablation}
\end{table}

\subsection{Hyper-parameter Analysis - Layer Number}

\begin{figure}
    \centering
    \includegraphics[width=0.65\linewidth]{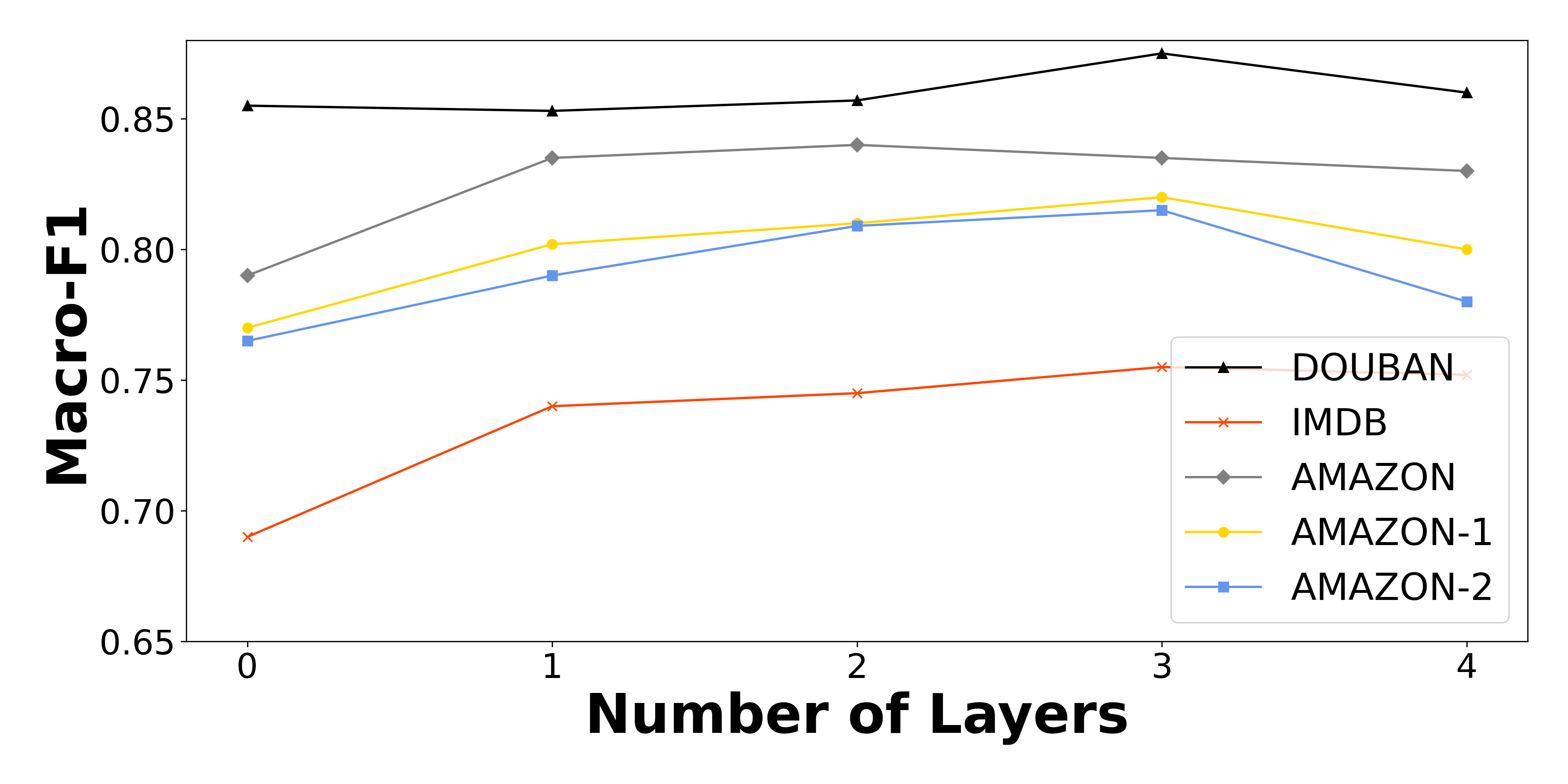}
    \caption{Macro-F1 values with varied layer numbers}
    \label{fig:layer}
\end{figure}

We change the layer number $K$ from 0 to 4, as shown in Figure~\ref{fig:layer}. It can be observed that the trend of Macro-F1 scores is similar across all datasets and arrives an optimal value when $K=3$. That is because adopting fewer than 3 layers prevents sufficient information exchange, whereas the continual increase of layers would lead to over-smoothing. 

\section{Conclusion}

This paper delves into the intricate problem of node representation learning within multi-modal heterogeneous networks, characterized with complicated interactions of modalities and node/edge types. 
To overcome the limitations associated with early or late fusion of multi-modal features, we put the fusion inside the GNN-based propagation process, thereby prompting node representations to align closely with category labels.
Notably, the innovative inter-modal attention acting on the modal-specific inter-node attention is proposed to enable adaptive modal fusion, based on the heterogeneous graph transformer framework. 
Moreover, another two critical factors in multi-modal data, modality alignment and missing, are also integrated into the model in a straightforward way to achieve significant improvements on node classification.

Future work will extend this method to more tasks demanding node representation learning with network structures.







\section*{Acknowledgements}
This work is supported by CAS Project for Young Scientists in Basic Research (YSBR-040), National Natural Science Foundation of China (62373061), Beijing Natural Science Foundation (L232028), and the Project of ISCAS (ISCAS-JCMS-202401).

\bibliographystyle{named}
\bibliography{sample-base}

\clearpage

\appendix

\section{Computational Complexity Analysis}
\label{sec:complexity}

In the heterogeneous network propagation module, the embedding of each node $v_i$ is updated once through aggregating the information from all of its neighbors with linear combination, in which two inter-node attention scores as aggregation weights are computed. 
The first one $\beta_{ij}^{(k)}$ in the cross-modal influence unit (Equations \ref{g}-\ref{beta}) handles modalities linearly, as it is independent of the current (influenced) modality $m$, so can be uniformly computed before the aggregation for each modality. 
The second weight $\bar{\beta}_{ij}^{(k)}$
in Equation \ref{tildebeta} for attention modulation considers each modality pair and is thus quadratic.
In addition, the pre-processing module, the feature fusion module and the loss functions are all obviously linear in terms of both the node number and the modality number.
Therefore, the worst-case complexity of the model is $O(|\mathcal{V}|^2 \cdot |\mathcal{M}|^2)$. More specifically, it can be $O(|\mathcal{V}| \cdot n \cdot |\mathcal{M}|^2)$, where  $n=\max_{v_i\in\mathcal{V}} N_i$ is the maximum number of neighbors for all nodes.
Notice that in most of multi-modal heterogeneous networks and multi-modal knowledge graphs, $N_i$ is much smaller than the node number $|\mathcal{V}|$, so the actual complexity is much lower than $O(|\mathcal{V}|^2 \cdot |\mathcal{M}|^2)$.



\section{Details of Baselines}
\label{sec:baseline-detail}
According to Section \ref{sec:baselines}, the baselines chosen in this paper involve three types:
\begin{enumerate}
    \item Typical heterogeneous graph neural networks without specially handling attributes of multiple modalities.
\begin{itemize}
\item \textbf{HAN} \cite{wang2019heterogeneous} proposes a novel heterogeneous graph neural network to generate node embeddings by aggregating features from meta-path based neighbors via hierarchical attentions.
\item \textbf{SHGP} \cite{yang2022selfsupervised} presents a self-supervised heterogeneous graph pre-training approach consisting of two modules with a shared attention-based aggregation mechanism, where the Att-LPA module generates pseudo-labels via structural clustering to guide the Att-HGNN module in learning attention scores and node embeddings.
\item \textbf{SeHGNN} \cite{sehgnn} introduces a simple and efficient heterogeneous graph neural network that reduces the complexity by using a lightweight mean aggregator for neighbor aggregation and a single-layer structure with long meta-paths.
\item \textbf{HERO} \cite{mo2024HERO} designs a self-supervised pre-training method to capture both homophily and heterogeneity from the subspace as well as nearby neighbors, discarding pre-defined meta-paths. 
\item \textbf{HGT} \cite{hu2020HeterogeneousGraphTransformer} proposes a heterogeneous graph transformer architecture by using node/edge-type dependent functions/parameters for attention-based aggregation, and employs relative temporal encoding to capture dynamic structural dependencies.
\end{itemize}
\item Special heterogeneous graph neural networks explicitly tackling attributes of multiple modalities.
\begin{itemize}
\item \textbf{HetGNN} \cite{zhang2019HetGNN} (by early fusion) 
utilizes Bi-LSTM to encode the feature of each modality, fuses them with mean pooling before intra-type and inter-type aggregations, and then learns node representations through a link-based self-supervised loss. 
\item \textbf{MHGAT} \cite{mhgat} (by late fusion) proposes a dual-level heterogeneous graph attention network for node-level and type-level aggregations to address graph heterogeneity, and uses the modality-level attention for final feature fusion.
    \begin{itemize}
      \item \textbf{MHGAT}-max realizes the node-level aggregation with the max aggregator.
      \item \textbf{MHGAT}-mean realizes the node-level aggregation with the mean aggregator.
    \end{itemize}
\item \textbf{XGEA} \cite{xu2023cross} integrates visual and semantic (textual) information into the GNN-based propagation process by leveraging features from one modality as complementary evidence to compute node similarity and attention scores of another modality.
\end{itemize}
\item Translation-based embedding methods treating the network structure as a new modality.
\begin{itemize}
\item \textbf{IDKG} \cite{li2023incorporating} obtains structural embeddings from the knowledge graph as an additional modality source of nodes, and integrates them with existing features of visual and textual modalities to constitute unified representations. 
\end{itemize}
\end{enumerate}

\begin{table}
   \small
\setlength{\tabcolsep}{2pt} 
\centering
\begin{tabular}{cccccc}
\toprule
\textbf{Variant} & \texttt{DOUBAN} & \texttt{IMDB} & \texttt{AMAZON} & \texttt{AMAZON-1} & \texttt{AMAZON-2} \\ 
\midrule
Text-only & 0.7624          & 0.6205     & 0.8002          & 0.6803  & 0.6803 \\ 
Vision-only &0.8459       & 0.7329     & 0.7999         & 0.6709          & 0.6709 \\  
\midrule
Node-ind &0.8580       & 0.7356     & 0.8326         & 0.7972          & 0.7707 \\ 
Edge-ind &0.8612       & 0.\textit{7536}    & \textit{0.8422}         & 0.8146          & 0.7277 \\ 
\midrule
Nonlinear (MLP)       & 0.8647          & 0.7335          & 0.8029          & 0.8193          & 0.8015          \\

\midrule
               \textbf{Ours} & \textbf{0.8758} & \textbf{0.7560} & \textbf{0.8427} & \textbf{0.8233} & \textit{0.8182}          \\ 
\bottomrule
\end{tabular}
  \caption{Macro-F1 values of additional ablative models. The best scores are marked in bold and the second-best in italic, counted together with Table \ref{tab:ablation}.}
  \label{tab:ablation2}
\end{table}

\section{Additional Ablation Study}
\label{sec:ablation2}

Besides the ablative models (in Section \ref{sec:ablation}) reflecting the core idea of mutual influence of modalities, we also examine other variants involving the fundamental elements in multi-modal heterogeneous networks.

\begin{figure*}
    \centering

        \begin{minipage}[b]{0.18\textwidth}
        \centering
        \includegraphics[width=\textwidth]{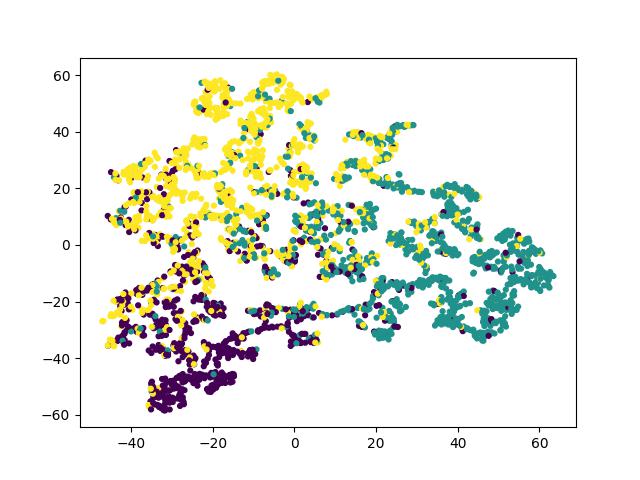}
        \caption*{(a) HAN}
    \end{minipage}
    \hspace{0.01\textwidth}
        \begin{minipage}[b]{0.18\textwidth}
        \centering
        \includegraphics[width=\textwidth]{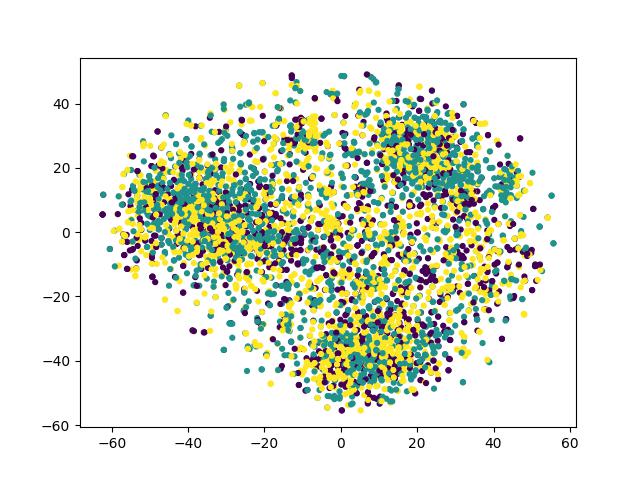}
        \caption*{(b) SHGP}
    \end{minipage}
    \hspace{0.01\textwidth}
    \begin{minipage}[b]{0.18\textwidth}
        \centering
        \includegraphics[width=\textwidth]{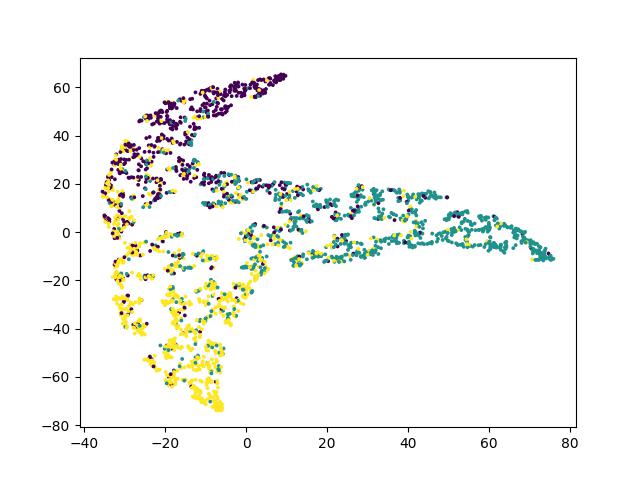}
        \caption*{(c) SeHGNN}
    \end{minipage}
    \hspace{0.01\textwidth}
        \begin{minipage}[b]{0.18\textwidth}
        \centering
        \includegraphics[width=\textwidth]{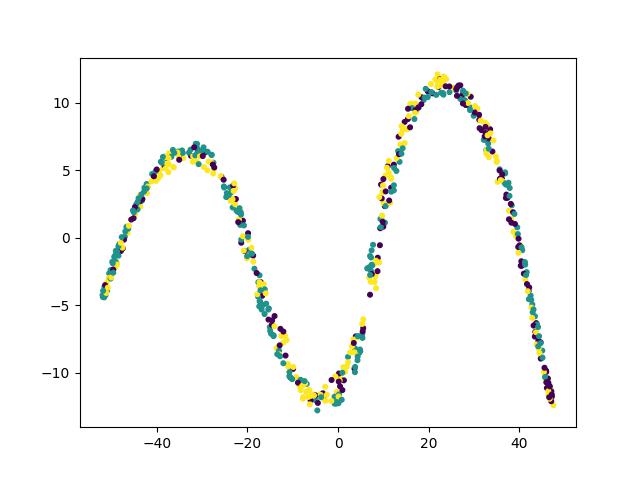}
        \caption*{(d) HERO}
    \end{minipage}
    \hspace{0.01\textwidth}
        \begin{minipage}[b]{0.18\textwidth}
        \centering
        \includegraphics[width=\textwidth]{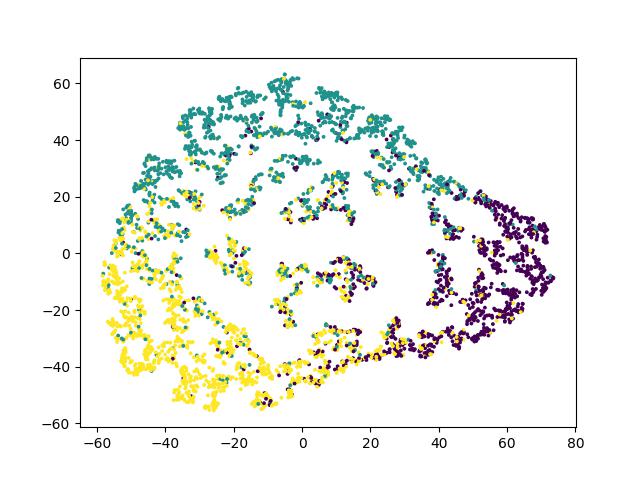}
        \caption*{(e) HGT}
    \end{minipage}
    \begin{minipage}[b]{0.18\textwidth}
        \centering
        \includegraphics[width=\textwidth]{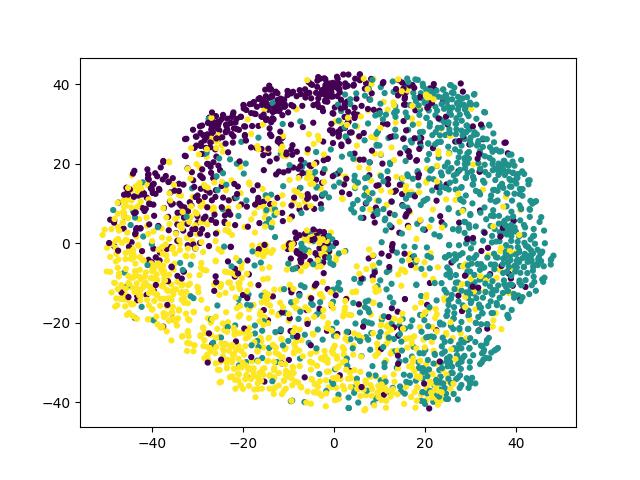}
        \caption*{(f) HetGNN}
    \end{minipage}
    \hspace{0.01\textwidth}
     \begin{minipage}[b]{0.18\textwidth}
        \centering
        \includegraphics[width=\textwidth]{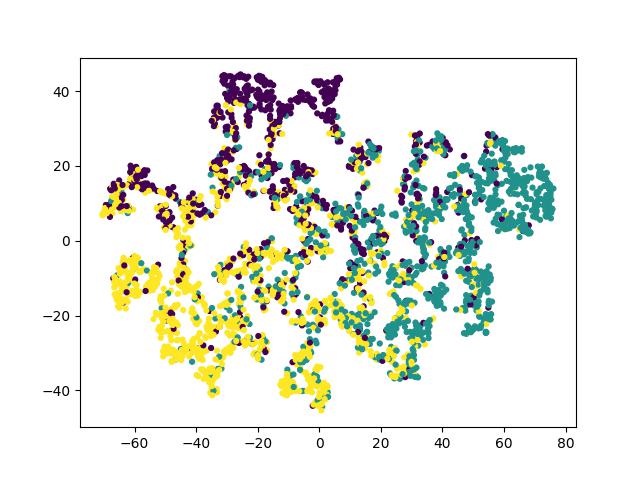}
        \caption*{(g) MHGAT}
    \end{minipage}
        \hspace{0.01\textwidth}
     \begin{minipage}[b]{0.18\textwidth}
        \centering
        \includegraphics[width=\textwidth]{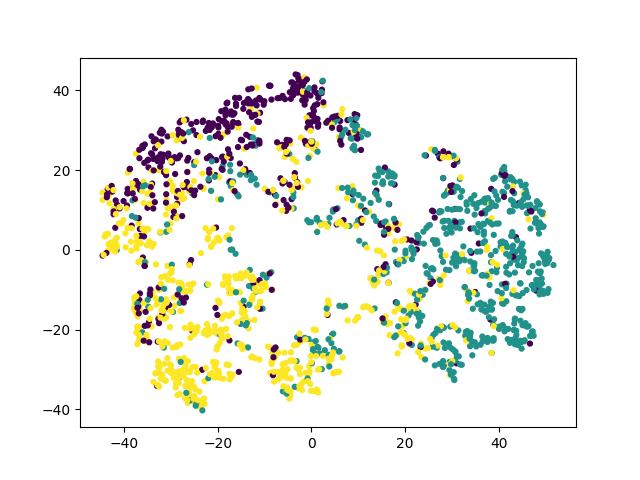}
        \caption*{(h) XGEA}
    \end{minipage}
        \hspace{0.01\textwidth}
         \begin{minipage}[b]{0.18\textwidth}
        \centering
        \includegraphics[width=\textwidth]{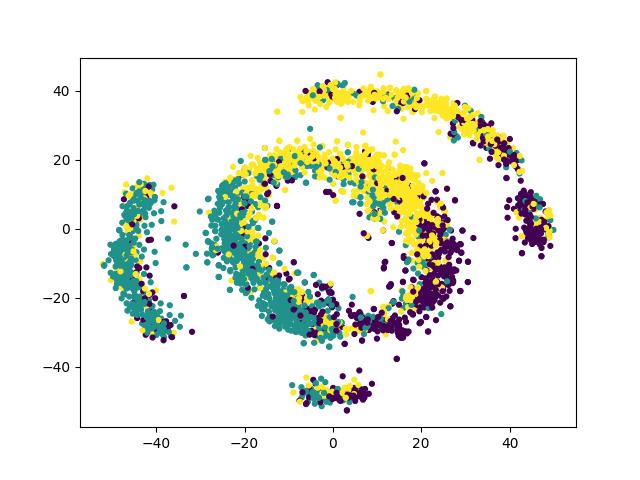}
        \caption*{(i) IDKG}
    \end{minipage}
    \hspace{0.01\textwidth}
    \begin{minipage}[b]{0.18\textwidth}
        \centering
        \includegraphics[width=\textwidth]{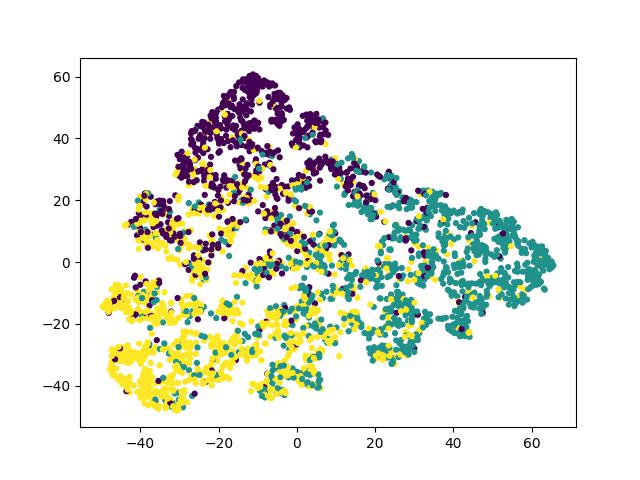}
        \caption*{(j) HGNN-IMA}
    \end{minipage}
    \caption{Visualization of node embeddings learned by HGNN-IMA and baselines on the \texttt{IMDB} dataset using t-SNE.}
    \label{fig:visualization}
\end{figure*}

\begin{itemize}
    \item 
    Not leveraging multi-modal information
    \begin{itemize}
        \item Text-only: It just utilizes textual attributes of nodes to learn node representations, without considering the mutual influence of modalities in propagation and the later fusion.
        \item Vision-only: It just utilizes visual attributes of nodes to learn node representations, without considering the mutual influence of modalities in propagation and the later fusion.
    \end{itemize}
    \item 
    Not leveraging heterogeneous information
    \begin{itemize}
        \item Node-ind: It does not distinguish the linear functions ($l^\textrm{I}$, $l^\textrm{K}$, $l^\textrm{Q}$, $l^\textrm{M}$ and $l^\textrm{A}$) for different node types, losing node heterogeneity.
        \item Edge-ind: It does not distinguish the matrices ($W^\textrm{NODE}$, $W^\textrm{MODAL}$ and $W^\textrm{MSG}$) for different edge types, losing edge heterogeneity.
    \end{itemize}
    \item 
    Complicating the linear transformations
    \begin{itemize}
   \item Nonlinear (MLP): It investigates an alternative architecture by replacing all the linear transformations in the model with nonlinear MLPs.
    \end{itemize}
\end{itemize}

As shown in Table \ref{tab:ablation2}, the advantage of our main model verifies the necessity of simultaneous utilization of multi-modal and heterogeneous information when categorizing nodes in MMHNs, to accommodate the complicated interactions among multiple modalities and node/edge types.

\section{Visualization on Embeddings and Attention Scores}
\label{sec:visualization}


\subsection{Learned Embeddings}
\label{sec:visual-embedding}
For an intuitive comparison, 
we employ t-SNE \cite{JMLR:v9:vandermaaten08a} to project the embeddings of target nodes into a 2-dimensional space. Taking the \texttt{IMDB} dataset for example, the embeddings of movie nodes in the test set are visualized in Figure \ref{fig:visualization}. Compared to all baselines, 
we observe that the embeddings learned by HGNN-IMA (Figure \ref{fig:visualization}(j)) exhibit higher intra-category similarity and can separate movies of different categories with more distinct boundaries.


\subsection{Attention Scores}
\label{sec:visual-attention}
To validate the effectiveness of the Cross-modal Influence Unit which embodies our core idea, we examine the relationship between the inter-node attention $\beta_{ij}$ computed in our model and the traditional inter-node attention $\alpha^m_{ij}$ without cross-modal influence used in the ablative model HGNN-IMA$-$cross. 
Notice that if the propagation process is consistent with the category labels, for the two nodes $v_i$ and $v_j$ belonging to the same category (called a positive pair), the attention score reflecting the propagation strength from $v_j$ to $v_i$ should be large, and vice versa for a negative pair.

\begin{figure}
    \centering
    \includegraphics[width = 0.7\columnwidth]{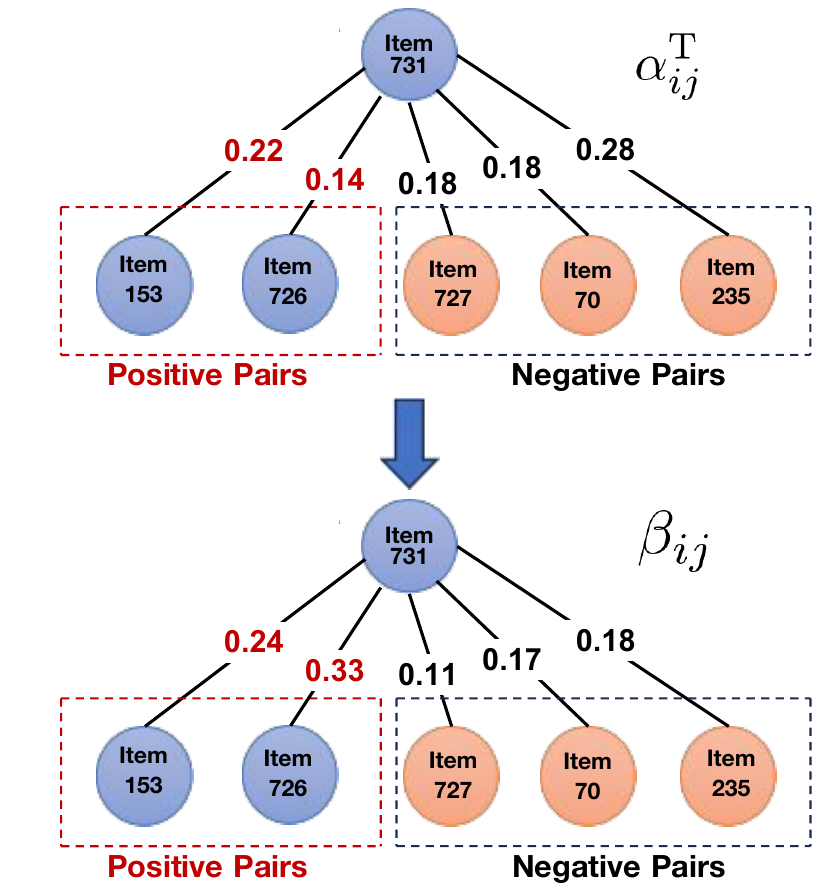}
    \caption{Case study on the comparison of inter-node attention scores $\alpha_{ij}^{\text{T}}$ in Cross-modal Influence Unit and traditional ones $\beta_{ij}$.}
    \label{fig:casestudy}
\end{figure}

As to the \texttt{AMAZON-1} dataset, we find compared to the traditional attention score $\alpha_{ij}^{(K),\text{T}}$ on the textual modality at the last layer, the modified $\beta_{ij}^{(K)}$ is larger for 54.5\% of positive node pairs with the same label, and smaller for 55.5\% of negative pairs.
Figure \ref{fig:casestudy} shows an example of node Item 731. We calculate the two attention scores $\beta_{ij}$ and $\alpha_{ij}^{\textrm{T}}$ between Item 731 and each of its neighbors respectively. Two of them form positive pairs and three of them form negative pairs. We can see that from $\alpha_{ij}^{\textrm{T}}$ to $\beta_{ij}$, the attention scores increase for both positive pairs and decrease for all negative pairs. Regarding the nodes of Item 153 and Item 726 in the former case, more information is exchanged with Item 731 during aggregation, so they would get similar embeddings and become easier to be assigned to the same category.  

\end{document}